*Title:* Embedding bifurcations into pneumatic artificial muscle

**Authors:**
N. Akashi,[1]* Y. Kuniyoshi,[2] T. Jo,[3] M. Nishida,[3] R. Sakurai,[3] Y. Wakao,[4] K. Nakajima[2]

**Affiliations:**
[1]Graduate School of Science, Kyoto University;
Kyoto, Japan

[2]Graduate School of Information Science and Technology, The University of Tokyo;
Tokyo, Japan.

[3]Digital Engineering Division, Bridgestone Corporation;
Tokyo, Japan.

[4]Advanced Materials Division, Bridgestone Corporation;
Tokyo, Japan.

*Email: akashi.nozomi.84z@st.kyoto-u.ac.jp

**Abstract:** Harnessing complex body dynamics has been a long-standing challenge in robotics. Soft body dynamics is a typical example of high complexity in interacting with the environment. An increasing number of studies have reported that these dynamics can be used as a computational resource. This includes the McKibben pneumatic artificial muscle, which is a typical soft actuator. This study demonstrated that various dynamics, including periodic and chaotic dynamics, could be embedded into the pneumatic artificial muscle, with the entire bifurcation structure using the framework of physical reservoir computing. These results suggest that dynamics that are not presented in training data could be embedded by using this capability of bifurcation embeddment. This implies that it is possible to embed various qualitatively different patterns into pneumatic artificial muscle by learning specific patterns, without the need to design and learn all patterns required for the purpose. Thus, this study sheds new light on a novel pathway to simplify the robotic devices and training of the control by reducing the external pattern generators and the amount and types of training data for the control.

**Main Text:**

**INTRODUCTION**

Recent studies have revealed that mechanical devices can be designed to use their body dynamics for desired information processing, such as a mechanical random number generator (*1*) and mechanical neural networks (*2*). Furthermore, the natural dynamics of mechanical bodies not designed for computation can be used as an information processing resource. The complex dynamics in soft robotic arms, which are inspired by the octopus, can be used for real-time computation, embedding a timer, and controlling the arm by employing the approach of physical reservoir computing (PRC) (*3–7*). Reservoir computing (*8–10*) is a recurrent neural network framework characterized by use of a high-dimensional neural network with nonlinearity and memory. In PRC (*11*), the network is replaced with physical dynamics. There are various types of robotic bodies, such as mechanical spring-mass-dampers (*12*), tensegrity (*13*), quadruped robots (*14*), and fish robots (*15*). This suggests that body dynamics can be directly exploited for information processing and control without external memory and nonlinearity.

A pneumatic artificial muscle (PAM) is a typical soft actuator that realizes expansion/contraction or bending dynamics through air pressurization. PAM been studied since the dawn of soft robotics (*16, 17*). The McKibben PAM (*18, 19*) is a central component of soft machines and devices, such as wearable devices (*20*) and robotic arms (*21*), and has the advantages of high durability against impact and vibration, a high force-to-weight ratio, and low manufacturing costs. In addition, PAM has been studied as a physical reservoir. PAM length sensors can be emulated by other sensory values in the PAM using the PRC framework (*22*). The air pressure of a rubber tube connected to a PAM that is attached to an assistive walking device could estimate the posture of the wearer using PRC (*23*). Moreover, control led to the assistant timing of the walking device based on the estimated information (*23*). Periodic dynamics have been embedded into a robotic arm composed of PAMs with PRC closed-loop control (*24*).

The bifurcation structure in dynamical systems is characterized by a qualitative change in dynamics, such as periodic and chaotic, through changes in a parameter. Bifurcation structures appear in the dynamics of robot bodies (*25, 26*). Bifurcation structures in central pattern generators have been shown to contribute to providing the capabilities of exploration and self-organized adaptation for robot control (*27–29*). Recently, it has been reported that artificial neural networks with rich information processing capabilities can reconstruct the entire bifurcation structure by learning a subset of dynamics included in the bifurcation structure, which we call bifurcation embedment (*30–34*). This study is the first attempt to realize bifurcation embedment into the robotic body. Embedding dynamics implies internalizing the central pattern generator into the body, which is usually externally attached to the robot. In addition, embedding bifurcation structures suggests that it is possible to control various qualitatively different patterns by learning several patterns, without the need to learn all patterns required in robot control. Such a generalization is more powerful than interpolation or extrapolation in traditional machine learning, as the properties of unseen dynamics in the bifurcation structure are qualitatively different from trained ones. Concretely, we demonstrated that periodic dynamics could be embedded into a PAM by training only chaotic dynamics, and vice versa. This study provides insight into reducing the weight and keeping the softness of robotic devices, such as wearable devices, by internalizing external pattern generators using computational capabilities in these morphologies. Furthermore, bifurcation embedment has the potential to significantly reduce the amount and types of training data for robot control.

## RESULTS
### Pneumatic artificial muscle
This study used the McKibben PAM (Fig. 1A), which consists of a cylindrical rubber tube covered by a braided cord. This PAM is forced by a nearly constant external load. If the PAM is pressurized, it expands in the radial direction and shrinks lengthwise. We used the PAM as a physical reservoir by injecting an input value as a control pressure and sensing its physical quantities. The measurement system is illustrated in Fig. 1B. Inner pressure, length, load, and electric resistance of the rubber were measured. Although traditional natural rubber has low conductivity, and it is difficult to measure its electric resistance value, this study increased the rubber conductivity from $1.0 \times 10^{-3}$ S/m to 20 S/m by mixing carbons with the rubber (*35*). The length of the PAM was 108 mm, with an outer diameter of 11 mm, an inner diameter of 9 mm, and an angle of braid $\pi/6$ rad in the equilibrium state (*35*).

## Dynamics

Fig. 1C shows the typical behaviors of each sensor value used in PRC. The input value $u(t)$ represents the following piecewise constant periodic signal:

$$u(t) = Au_n + B \left(u = \left[\frac{t}{\tau}\right]\right), \tag{1}$$

$$u_n = \sin\frac{2\pi\tau}{T}n, \tag{2}$$

where $A$ and $B$ are the input magnitude and bias, respectively, which tune the input to a suitable input range for the device. $\tau$ is the input interval, and $T$ is the period of input. This study used the following input parameters: A = 0.2 MPa, B, = 0.3 MPa, $\tau$ = 0.1 sec, and $T$ = 1.2 sec. Sensor responses are presented in Fig. 1C. Sensor value sampling timing was the timing of updating the next input signal. Measured pressure showed nearly the same behavior as control pressure. Furthermore, length and resistance exhibited different curves between the compression and extension phases and possessed the nonlinearity and hysteresis for the input. The load value barely responded to the control pressure and was likely a noise signal.

## Bifurcations of electric resistance through external load

The behavior of the electric resistance of the rubber changed drastically through external load and explained the mechanics of these bifurcations. This was accomplished by focusing on changes in the rubber's thickness. The external load added to the PAM changed by 5 N from 100 N to 250 N under the periodic input signal, as depicted in Fig. 1C and represented by Eq. 1. The relationship between length and resistance is presented in Fig. 2A. Resistance response was opposed after 106 mm, which is approximately the equilibrium length of the PAM ($d_0 = 108$ mm). Thus, these regions can be considered to correspond to contraction and expansion phases. The length gap between 106 mm and 108 mm is considered to occur due to offset measurements. The electric resistance of a rubber tube has the same tendency as the rubber thickness (*36*). Rubber thickness was derived from measurements of diameter and conservation of volume. The thickness model is provided in the Materials and Methods section. The thickness of the rubber peaks at the equilibrium length is depicted in Fig. 2A. In the expansion phase, the thickness $d_e$ thinned out from the equilibrium thickness $d_0$ because the inner diameter did not change. In the contraction phase, the thickness $d_c$ thinned out from $d_0$ due to expansion in the radial direction. Therefore, resistance peaks at the equilibrium length as rubber thickness.

Fig. 2B illustrates the response of length for applied load and pressure in the experiment. Using this method, three load regions were identified: compression phase alone, compression and extension mixing phase, and extension phase alone. The typical resistance time series of each phase is presented in Fig. 2C. Resistance in the compression phase responded to anti-phase pressure value. Resistance in the mixing phase changed to a two-peak behavior. In contrast to the compression phase, resistance in the extension phase responded to the pressure value in-phase. Fig. 2D depicts the bifurcation diagram, where local minimum values of resistance are plotted for the applied load values. Bifurcations were confirmed, revealing that the local minimum value changed from one to two at 160 N and became one again at 220 N. These bifurcation points correspond to change points of the compressing, mixing, and extension phases.

**Computing scheme**

Fig. 3 illustrates the computational scheme of PAM PRC. The input signal was injected as a control pressure, which is a one-dimensional value. The PAM acted as a physical reservoir by providing a nonlinear historical response to the input. We obtained reservoir variables by sensing these responses and constructing output values from a weighted sum of reservoir variables and bias term.

The input signal is a piecewise constant one-dimensional signal, which is represented in Eq. 1. The nonlinearity and memory of the physical reservoir can be tuned by tuning the input magnitude $A$ and input interval $\tau$ in Eq. 1. The dependency of nonlinearity and PAM physical reservoir memory on input magnitude $A$ and input interval $\tau$ is discussed in the Supplementary Materials 2. Sensor values at time $t$ are represented in the form $s(t) = (s_1(t), \ldots, s_M(t))$, where $M$ is the number of sensor values used as reservoir variables. We obtained reservoir variable $x_n$, which corresponds to input $u_n$, based on sensing $L$ times from input injected time $t$ to input updated time $t + \tau$. Eq. 2 presents $x_n \in \mathbb{R}^{ML+1}$:

$$x_n = \left[ s(t); s\left(t + \tau \frac{1}{L}\right); \cdots; s\left(t + \tau \frac{L-1}{L}\right); 1 \right], \tag{3}$$

where the number of samples is represented by $L$. This multiplexing method (37), referred to as time-multiplexing, boosts the computational power of the reservoir from a small number of variables and has been widely used in PRC (6, 38, 39). In this study, $L$ was fixed at five in all the experiments. The output values $\hat{y}_n = (\hat{y}_{n,1}, \ldots, \hat{y}_{n,K}) \in \mathbb{R}^K$ were generated as follows:

$$\hat{y}_n = W_{out}^\top x_n, \tag{4}$$

where $W_{out} \in \mathbb{R}^{(ML+1)K}$ is output weight (+1 in the index means a bias term). The output weight is obtained by ridge regression:

$$W_{out} = (X^\top X + \lambda I)^{-1} X^\top Y, \tag{5}$$

where $X = (x_{N_{wash}+1} \cdots x_{N_{wash}+N_{train}}) \in \mathbb{R}^{(ML+1)N_{train}}$ is the training data matrix, $Y = (y_{N_{wash}+1} \cdots y_{N_{wash}+N_{train}}) \in \mathbb{R}^{KN_{train}}$ is a target data matrix corresponding, and $\lambda$ is the ridge parameter. The numbers of washout and training data are represented as $N_{\text{wash}}$ and $N_{\text{train}}$, respectively.

The open- and closed-loop settings are depicted in Fig. 3. Open-loop represents a case in which the input signal $u_n$ is external to the reservoir. Closed-loop represents the case in which the input signal $u_n$ is the output value of the reservoir one time step prior. In the closed-loop setting, input values, which are control pressures, are restricted to a certain range $[u_{\min}, u_{\max}]$ to prevent the breakdown of the system.

$$u_n = \begin{cases} u_{\min} & (\hat{y}_{n-1} < u_{\min}) \\ \hat{y}_{n-1} & (u_{\min} \leq \hat{y}_{n-1} \leq u_{\max}) \\ u_{\max} & (\hat{y}_{n-1} > u_{\min}) \end{cases} \tag{6}$$

This study used the value $(u_{\min}, u_{\max}) = (0 \text{ MPa}, 0.5 \text{ MPa})$.

**Information processing capacity of a single pneumatic artificial muscle**

Information-processing capabilities, that is, the nonlinearity and memory of each sensor value in the PAM, were revealed. We used the *information processing capacity* (IPC) (*40*) criteria, which describes the function that dynamical system serves for an input signal from independent and identically distributed (i.i.d.) random variables (*41*). Memory and linear/nonlinear transformation capability of the reservoir can be obtained by checking the bases of this function. IPC limited linear components are called *memory capacity* (*42*). Detailed definitions and formulation of the IPC are provided in the Supplementary Materials 1. The IPC restricted the delay to $\leq D$ and degree of polynomial functions to $\leq K$, as $\text{IPC}[D, K]$. $\text{IPC}[D, K]$ can be decomposed by function reconstruction capacities $C[\mathcal{Y}]$, where $\mathcal{Y}$ is an orthogonal basis function in the focusing functional space. The $\text{IPC}[1, K]$ is referred to as *memory capacity*. For a number $N$ of the linear independent states of reservoir variable, the following theoretical equation holds (40):

$$\lim_{D,K \to \infty} IPC[D, K] \leq N \tag{7}$$

where equality is established if the reservoir has *echo state property* (ESP). Here, ESP, which is an important property for reservoir computing, guarantees the reproducibility of the computing results (*43*).

First, we showed the IPCs of each single sensor value and multiplexing sensor values in the PAM, which included pressure, length, resistance, load, and all sensors combined and time-multiplexed. The IPCs are presented in Table 1. The IPC of the pressure was nearly one, and the pressure could be completely described by the input sequence. Therefore, the pressure could be a computational node that has the ESP for input. The IPCs of length and resistance were slightly lower than one, and these sensor values could nearly be described by the input sequence; however, they had few irreproducible components. The IPC of the load was nearly zero, and load moved nearly independently of the input. The IPC could be improved to approximately 10 by combining all types of sensors and using time-multiplexing. The total number of reservoir variables was $20 = 4 \times 5$. An IPC lower than the number of variables implied the existence of input-independent or linearly dependent components.

Further details, such as nonlinear and memory capacities of the IPCs in the PAM, were examined. Fig. 4 depicts the dependency of the IPCs on the external load. The capacities of the pressure did not change through external load. The capacities of delay zero in the length monotonically increased as external load increased from 50 N to 250 N. Therefore, a PAM with a smaller external load could produce information processing that requires more memory. However, the degree components in the resistance non-monotonically changed through the external load. In addition, the linear components monotonically increased as external load increased to 150 N, degree two components were dominant when the external load was 175 N and 200 N, and the linear components were dominant when the external load was greater than 225 N. Increasing and decreasing switches corresponded to the bifurcation points of resistance; therefore, these critical behaviors were derived from the intrinsic resistance bifurcations.

**Open-loop experiments**

We evaluated the performance of PAM PRC in an open-loop setting. We investigated PAM length sensor emulation (*35*), which is a practical task. PAM length sensor emulation is a real-world task that emulates the PAM length sensor value from the input pressure value. A laser displacement sensor, which is a standard length sensor for the PAM, is made of a rigid component that takes away the softness of the PAM. Therefore, it is a practically important method to ensure the softness of the PAM, emulate the length sensor by other sensory values, eliminate it from the PAM. Although the length dynamics of the PAM respond nonlinearly to hysteresis for the input pressure, PAM length time series can be predicted by a recurrent neural network (*35, 44–48*). Here, the input sequence in this task arose from uniformly random values and was transformed to $[0, 0.5]$ MPa control pressure value, as $(A, B) = (0.5, 0)$ in Eq. 1. In addition, we provided the performance of PAM PRC for a typical benchmark task in the Supplementary Materials 3.

Furthermore, we evaluated the performance of the task using the normalized mean squared error (NMSE), as follows:

$$\text{NMSE} = \frac{\frac{1}{N_{\text{eval}}} \sum_{n=1}^{N_{\text{eval}}} (\hat{y}_n - y_n)^2}{\sigma^2(y_n)} \tag{8}$$

where $N_{\text{eval}}$ is the number of evaluation data. In the following experiments, the number of washout data was $N_{\text{wash}} = 1{,}000$, the number of training data was $N_{\text{train}} = 40{,}000$, and the number of evaluation data was $N_{\text{eval}} = 9{,}000$. We compared the performances between PAM PRC and echo state network (ENS) (*9*), which is a typical recurrent neural network in reservoir computing. The ENS had the same number of computational nodes as the number of reservoir variables in PAM PRC. The best hyperparameters of the ENS for each task were determined by the grid search. In addition, we compared the performance of the physical model in the length sensor emulation task. The formulations of the ESN and physical model are presented in the Materials and Methods section.

We multiplexed sensor values in the PAM using time-multiplexing $L = 5$. In one instance, we used 15 reservoir variables (which are time multiplexed pressure, resistance, and load), whereas in another, we used 10 reservoir variables, which were time multiplexed pressure and resistance. The number of nodes in the ESN was the same as in the case of using the load, so this is 15.

The NMSEs of the physical model, ESN, PAM PRC without loads, and PAM PRC with loads were 0.0893, 0.0436, 0.0302, and 0.0294, respectively. PAM PRC outperformed the ESN and physical model. Fig. 4A illustrates the time series of the input, reservoir variable, target, and output signals. The responses of pressure and resistance time series were induced by the random input sequence; however, the load time series was independent of the inputs, as was the result of the IPCs for the load signals in Table 1. Although the load was not induced into the input signal information, cases using load signals had a higher performance than those without the load. In addition, the performance using the load was higher than the upper bound of the performance of the ENS with the sufficient number of nodes, as depicted in Fig. 4B. The length value was not only driven by the input signal but also reflected the noise or intrinsic time-varying dynamics, as the IPC length was 0.975. Only the part of the dynamics that was driven by input can be reconstructed by external machine learning, such as an ESN, which transforms only the inputs. Conversely, PAM PRC with load could treat not only explicit input signals but also implicit inputs and the internal state of the PAM, as the load was independent of the input due to the load IPC (Table 1).

There is another advantage of PAM PRC in the case of few training data. Fig. 4C present performance comparison between an ESN with 600 nodes and PAM PRC through training data. The performance of the ESN significantly decreased when the training data were less than 1,000; however, PAM PRC archives nearly the same performance when the number of training data ranged from 100 to 10,000. This could be considered a problem derived from the dimension of machine-learning networks. An ESN with 600 nodes produces overfitting when the training data are limited; however, PAM PRC can work with as few computational nodes such as 15. This does not easily produce overfitting, even with a small amount of data. Therefore, selecting a suitable reservoir for the task has advantages not only in performance and calculation costs but also with regard to learning.

**Closed-loop experiments: Attractor embedding**

This study analyzed closed-loop control by PRC in a single PAM. First, we analyzed the potential to embed attractors in a PAM. We focused on limit cycle, strange attractor of the logistic map, and the Rössler attractor. The limit cycle defined by Eq. 2 was a one-dimensional periodic dynamic. Such rhythm dynamics are important as a central pattern generator in robot control (*49*). In addition, not only periodic but also chaotic oscillators play an important role in robot control. As a central pattern generator, a chaotic oscillator can derive adaptive and exploratory behaviors by its complex dynamics (*28, 29*). The logistic map, which is a one-dimensional dynamical system with discrete time, was defined by the following equation:

$$y_{n+1} = ay_n(1 - y_n) \tag{9}$$

where $a$ is a model parameter and is set as a chaotic parameter $a = 3.7$. The embedding of logistic dynamics does not require memory, as the next step of the logistic map can be determined only by the current step of the logistic map. Furthermore, the Rössler attractor (*50*) of three-dimensional chaos with continuous time is defined by the following equation:

$$\begin{aligned} \dot{y}_1 &= -y_2 - y_3 \\ \dot{y}_2 &= y_1 + ay_2 \\ \dot{y}_3 &= b + y_1 y_3 - cy_3 \end{aligned} \quad (10)$$

where $a$, $b$, and $c$ are model parameters and set as typical chaotic parameters: $(a, b, c) = (0.2, 0.2, 5.7)$. As chaos with continuous time does not occur unless it is a dynamical system with at least three dimensions (*51*), and the nonlinear term, which is essential for chaos, is only $y_1 y_3$, this model can be considered one of the simplest models in chaos with a continuous time.

In the training phase, we injected $y_n$ into the reservoir using an open-loop and trained the output weight using $y_{n+1}$ as a teacher signal (this training scheme is referred to as *teacher forcing* (*52*)). The input range of all experiments was set to $[0.1, 0.5]$ MPa by tuning $A$ and $B$ in Eq. 1. The input signal of the Rössler system, which is a three-dimensional system, was $y_1$, which was discretized by a sampling interval of 0.5 sec. Although the input is one-dimensional, the attractor can be reconstructed with all dimensions due to Takens' embedding theorem (*53*) if the reservoir has sufficient memory and nonlinearity. Furthermore, the input interval of the PAM control pressure is $\tau = 0.1$ sec in the experiments with the limit cycle and Rössler system, and $\tau = 0.2$ sec in the experiments with the logistic map. The embedding of the limit cycle and Rössler system require memory; however, the embedding of the logistic map did not (the relationship between the input interval and IPC of PAM PRC is discussed in the Supplementary Materials 4). In all of the experiments, we fixed the number of washout data at $N_{\text{wash}} = 1,000$ and the number of training data at $N_{\text{train}} = 4,000$. In the prediction phase, we switched the open-loop and closed-loop after 1,000 time steps. In addition, we evaluate the embedding result using the output time series, the attractor in the delayed coordinate system, and power spectra.

Fig. 6 presents the results of the attractor embedding. In limit cycle embedding, the embedded attractor had similar properties as a time series attractor, and spectra as the target limit cycle. In logistic map embedding, the output time series deviated from the target signal as time passed after switching from an open- to a closed-loop due to the initial state sensitivity of chaos. However, there was an output signal close to the target attractor, and the embedding attractor captured the characteristics of the target attractor. In addition, the output signal had a broad range of spectra that was the same as the target signal, rather than a clear perk of spectra such as the limit cycle. Finally, in the Rössler attractor embedding, the output attractor could reconstruct the highest peak spectra of the Rössler attractor; however, it could not reconstruct multiple peaks, which are characteristic of chaos, that is, of the Rössler attractor.

Next, we evaluated the robustness of the attractor embedding. For this, we injected a random signal from the target signal and confirmed that the output signal could quickly return to the target attractor after switching to a closed loop. We focused on the limit cycle as a target attractor, and the input signals in the open-loop were zero and random signals. The results are depicted in Figs. 6D and 6E. The output signals quickly returned to the target attractor after switching.

**Closed-loop experiments: Bifurcation embedding**

We confirmed that the IPC of the resistance in the PAM could drastically change through the change in external load in an open-loop setting. Moreover, we found the change in the output signal of PAM PRC through the external load in a closed-loop setting. The following training data were used:

A) A limit cycle with a period of 1.2 sec with an external load of 100 N (same as the limit cycle in Fig. 6);
B) Limit cycles with periods of 1.2 and 2.4 sec with external loads of 100 N and 250 N, respectively;
C) The chaotic trajectory of the logistic map, where $a = 3.7$, with an external load of 100 N (same as in the case of the logistic map in Fig. 6);
D) The period 4 trajectory of the logistic map, where $a = 3.55$, with an external load of 100 N.

We confirmed the change in the output signal in closed-loop control when the external load changed by 5 N from 100 N to 250 N every at 2,000 time steps.

Fig. 7 depicts the results. In experiment A, the amplitude and frequency of the limit cycle continuously changed in the range from load 100 N to 200 N. However, the limit cycle structure of the output signal suddenly collapsed at an external load of 200 N, and the output signal changed to nearly static dynamics. This switching point was around the second bifurcation point of the resistance, as shown in Fig. 2D. Thus, dynamics may switch, as the bifurcation of the resistance propagated to the entire dynamics of the PAM via closed-loop control. Conversely, the result of experiment B indicated that it is possible to suppress the closed-loop bifurcations. In experiment B, we trained limit cycles with different frequencies when external loads were 100 N and 250 N. The results revealed that the frequency of the closed-loop dynamics with intermediate external loads was linearly interpolated.

The results of experiments C and D revealed that periodic and chaotic dynamics could be embedded simultaneously and that one of the dynamics could be generated from learning another one of them. In experiment C, we trained chaotic dynamics in the logistic map when the external load was 100 N. The dynamics switched from chaotic dynamics to period 2 dynamics when the external load was 170 N, which corresponded to the first step of the bifurcation of resistance. As the bifurcation diagram indicates, period 2 dynamics appeared intermittently, acting as a window of the period-doubling bifurcation. In experiment D, we trained period 4 dynamics in the logistic map when the external load was 100 N. The dynamics switched to chaotic dynamics with a one-dimensional attractor in the delay coordinate and broad spectra at the external load 200 N, which corresponded to the second bifurcation of the resistance. The chaotic attractor in the delay coordinate had an alternative shape, similar to a cubic function, to the logistic map. In addition, dynamics had an unstable fixed point near $y_{t+1} = y_t$, as there was a hole at the intersection of $y_{t+1} = y_t$ and the attractor.

These bifurcation embedding results could be useful for robotics applications. For instance, the automatic switching conducted in experiment A could be used for an emergency stop when the external load exceeds the threshold and an idling stop that transitions to a stationary state while the main power is on. This presents the possibility of internalizing adaptive behavioral control that depend on changes in the environment. In addition, the results of experiment B revealed that this switching can be turned off by explicit training on both sides of the bifurcation of the inherent dynamics. Furthermore, the results of experiments C and D demonstrated that multiple qualitatively different dynamics, including chaos, could be switched according to changes in the environment. The results of experiments C and D did not indicate the desired bifurcation structure but rather a bifurcation structure based on training data and reservoir dynamics. We present the embedding a bifurcation structure, which includes desired qualitative different signals, by training dynamics on both sides of bifurcation explicitly in the Supplementary Materials 4.

**DISCUSSION**

This study demonstrated that various dynamics and bifurcation structures can be embedded into a soft robotic actuator through systematic analyses of PAM PRC. These results reveal potentials, limitations, and future directions of computing using the robot body.

In the open-loop setting, PAM PRC can outperform the external ESN with a sufficient number of computational nodes by using resistance and load sensor values. This performance is believed to derive from the fact that the resistance and load sensor values reflect the PAM internal state, such as time-variant components and extra inputs, which cannot be represented by the pressure input. The evaluation of PAM IPCs should be extended to multi-inputs and time-variant form to test this hypothesis (*41*). External machine-learning networks can achieve the same level of performance as PAM PRC if resistance and load sensor values are injected into them. In addition, this study revealed that PAM PRC can obtain high performance of the sensor emulation from small-size training data. This aspect is an important advantage in the information processing of soft materials, as soft materials generally have lower durability than rigid materials, and their material properties can be easily changed over a long time period.

Furthermore, in the closed-loop control, we succeeded in embedding the attractor of the sinusoidal wave and logistic map but failed to embed the Rössler attractor. The dimension of the attractor is believed to have caused this failure. The sinusoidal wave and logistic map have one-dimensional attractor; however, the Rössler attractor that is embedded in a three-dimensional space has a fractal dimension ranging from two to three. The number of inputs in this study is one as the control pressure. To embed high-dimensional dynamics, such as Rössler, it is necessary to utilize the memory of the reservoir due to the Takens delay embedding theorem (53). Therefore, the failure could be because the memory of a single PAM is insufficient to reconstruct the target variables that are not injected as inputs.

In addition, we demonstrated that the PAM can be embedded qualitatively different attractors from the training attractor in the closed-loop experiments. These results suggest that bifurcations in the morphology may have a potential to be exploited to embed the bifurcation structure of the target dynamical system. However, the mechanism to embed the bifurcation structure into the reservoir has not been fully understood to date (33). Moreover, the necessity of the intrinsic bifurcations of the reservoir for the bifurcation embedding remains unknown.

This bifurcation embedding into the body suggests the strong potential of the control of robotics. For instance, if we can embed the period-doubling bifurcation in the morphology of the robot, it may be possible for the robot to generate all the arbitrary periodic dynamics and chaos underlying the Li-Yorke chaos (54) from learning only finite period patterns.

This study indicated that the body dynamics of the single PAM have high computational capability. We believe that these results can be expanded to practical situations. The structures that consist of multiple PAMs, such as a robot arm and wearable assistance suit, may have the potential to embed higher-dimensional and more complex dynamics than single PAMs. For example, if we can embed chaotic itineracy in the robot body, the robot can switch many primitive patterns autonomy and stochastically (55). Moreover, the embedded bifurcation structure could serve as an adaptive pattern switch for the environment, such as an anomaly detection and failure prevention of robot, as the bifurcation points correspond to the change points of the dynamic phase of the body dynamics, such as contraction and extension phases in the PAM.

## MATERIALS AND METHODS
### Pneumatic artificial muscle thickness model
The thickness model of the PAM is presented in Fig. 2A. Thickness was calculated using the following equation:
$$d = R - r \tag{11}$$
where we assume that the rubber tube in the PAM is a uniform cylinder and that $R$ and $r$ are the outer and inner radius of the cylinder, respectively. Furthermore, we assumed the below linear relationship between length and outer radius, as the coefficient of determination between the length and outer radius was 0.9934, which was obtained from the 9 values of length and thickness of the rubber tube with an external load of 50 N.
$$R = -0.3382l + 47.525 \tag{12}$$

where the length of the rubber tube is represented by $l$. We obtained the inner radius $r$ using the following equation due to the constraint of the constant volume of rubber and restriction of both ends of the tube:

$$r = \begin{cases} r_0 & (l \geq l_0) \\ \sqrt{R^2 - \dfrac{V}{l\pi}} & (l < l_0) \end{cases} \quad (13)$$

where $V$ and $S$ are the volume and cross-sectional area of rubber, respectively, and $r_0$ and $l_0$ are the inner radius and length in equilibrium length, respectively. Fig. 2A presents the length and thickness that were calculated using the above equations.

**Echo state network**

The architectures of the ESN were compared with PAM PRC. The $i$th computational node at time $t$ is represented as $x_t^i$, the $j$th input node is represented as $u^j$, and the $l$th output node at time $t$ is represented as $\hat{y}_t^l$. The computational nodes and outputs of the ESN are given by

$$x_t^i = f\left(A_{\text{cp}} \sum_{j=1}^N w_{i,j} x_{t-1}^j + A_{\text{in}} \sum_{j=1}^N w_{i,j}^{\text{in}} u_t^j\right) \quad (14)$$

$$\hat{y}_t^l = \sum_{j=1}^N w_{i,j}^{\text{out}} x_t^j \quad (15)$$

where the activation function is given by $f$, which is the hyperbolic tangent; each node of the input weight $W_{\text{in}} = \left(w_{i,j}^{\text{in}}\right)$ comprises a uniform distribution with $[-1, 1]$, and each node of the internal weight $W = (w_{i,j})$ comprises a uniform distribution with $[-1, 1]$ and is normalized to make the spectral radius one. The coupling magnitude $A_{\text{cp}}$ coincides with the spectral radius of $A_{\text{cp}}W$. The bias term $x_t^0$ is set as $x_t^0 = 1$. The output weight $W_{\text{out}} = (w_{out})$ is tuned by training. We fix $A_{\text{in}} = 1$ and optimize $A_{\text{cp}}$ by grid search for the range $[0, 1.2]$ in each task.

**Dynamical model of the PAM**

We estimated the length dynamics of PAM from the injected pressure and load for the control. The models of PAM dynamics have been widely investigated in previous studies (56,57). Based on these studies, we used the following length model of the PAM:

$$M\ddot{x} = -F_{\text{elas}}(x) - F_{\text{fric}}(\dot{x}) - F_{\text{pre}}(x, p(t)) + F_{\text{ex}}(t) \quad (16)$$

where the displacement of the PAM length is represented as $x$, and the mass of the PAM is represented as $M$; the elastic force of the rubber, friction of the rubber, and tension of volume change by pressure are represented as $F_{\text{elas}}(x)$, $F_{\text{fric}}(\dot{x})$, and $F_{\text{pre}}(x, p(t))$, respectively; and the input pressure and input load are represented as $p(t)$ and $F_{\text{ex}}(t)$. Here, tension by pressure is derived from the following Schulze equation:

$$F_{\text{pre}}(x, p(t)) = \frac{\pi D_0 p(t)}{4} \frac{1}{\sin \theta_0} (3(1-\varepsilon)^2 \cos \theta_0 - 1) \quad (17)$$

where the strain of the PAM is represented as $\varepsilon = (l_0 - x)/l_0$ and the equilibrium length, inner radius, and angle of the braided code are represented as $l_0, D_0$, and $\theta_0$, respectively. Note that the Schulze equation assumes that the PAM is a uniform cylinder with zero thickness. When the real PAM is compressed, it is not a cylinder but a bent shape, as both ends of the PAM are fixed. The model that considers the non-uniform and bent shape of the PAM has been previously proposed in extant literature (*57*). We ensured the linear elasticity $F_{\text{elas}}(x) \propto x$ is the elasticity of the PAM. We could accurately estimate the length of the PAM by solving the equation of the equilibrium of $F_{\text{pre}}(x)$, $F_{\text{elas}}(x)$, and $F_{\text{ex}}$ in the static state. However, in the dynamic state, in which the PAM continues to move, it is difficult to estimate the PAM dynamic, as the Schulze equation cannot consider the hysteresis depicted in Fig. 2 in the main text. The causes of the hysteresis can be considered as the effects of Coulomb and viscous frictions (*58–60*). Therefore, Eq. 25 can be rewritten using the following equation:

$$\ddot{x} = -Ax - B\dot{x} - C\text{sgn}(\dot{x}) + D\left(-F_{\text{pre}}(x, p(t)) + F_{\text{ex}}(t)\right) \tag{18}$$

Here, $A$, $B$, $C$, and $D$ are the parameters of the model. We optimized these parameters using grid search, and the parameters used in the Section 2.6 were $(A, B, C, D) = (6353, 80.05, 10, 0.635)$. The measured length values were offset due to a measurement error; thus, we added a bias to the length-predicting value from the physical model to coincide with the average values of the measured and predicted lengths.

## Supplementary Materials

All Supplementary Materials are available at:
https://figshare.com/s/abdd59978c30f24f7c0a
(This repository is currently set in private mode (no doi number), but available from the link.)

Supplementary Material 1. Detailed calculation method of information processing capabilities.
Supplementary Material 2. Dependencies of information processing capacities on controller, body, and environmental conditions.
Supplementary Material 3. Nonlinear autoregressive moving average task.
Supplementary Material 4. Embedding periodic and chaotic attractors into the same weight.
Fig. S1. Information processing capacities through input interval.
Fig. S2. Information processing capacities through input magnitude.
Fig. S3. Information processing capacities through equilibrium length.
Fig. S4. Information processing capacities through temperature.
Fig. S5. Time series of the target and output signals in the NARMA2 task.
Fig. S6. The results of the closed-loop during the chaos and period dynamics embedding.
Table S1. Base experiment conditions.
Table S2. Normalized mean squared errors in NARMA2 task.
Supplementary Data. PAM sensory time series for IPC calculation.
Movies S1. Open-loop length sensor emulation.
Movies S2. Closed-loop attractor embedding.
Movies S3. Closed-loop bifurcation embedding.

## References and Notes

1. N. Akashi, K. Nakajima, M. Shibayama, Y. Kuniyoshi, A mechanical true random number generator. *New J. Phys.* **24**, 013019 (2022).

2. R. H. Lee, E. A. B. Mulder, J. B. Hopkins, Mechanical neural networks: Architected materials that learn behaviors. *Sci. Robot.* **7**, eabq7278 (2022).
3. K. Nakajima, H. Hauser, R. Kang, E. Guglielmino, D. G. Caldwell, and R. Pfeifer, Computing with a muscular-hydrostat system, in *2013 IEEE International Conference on Robotics and Automation* (2013), pp. 1504–1511.
4. K. Nakajima, H. Hauser, R. Kang, E. Guglielmino, D. G. Caldwell, and R. Pfeifer, A soft body as a reservoir: case studies in a dynamic model of octopus- inspired soft robotic arm. *Frontiers in Computational Neuroscience. Front. Comput. Neurosci.* **7** (2013).
5. K. Nakajima, H. Hauser, T. Li, R. Pfeifer, Information processing via physical soft body. *Sci. Rep.* **5**, 10487 (2015).
6. K. Nakajima, H. Hauser, T. Li, R. Pfeifer, Exploiting the dynamics of soft materials for machine learning. *Soft Robot.* **5**, 339 (2018).
7. K. Nakajima, T. Li, N. Akashi, in *Robotic Systems and Autonomous Platforms*, S. M. Walsh, M. S. Strano, Eds. (Woodhead Publishing, 2019), pp. 181–196.
8. W. Maass, T. Natschlager, H. Markram, Real-time computing without stable states: A new framework for neural computation based on perturbations. *Neural Comput.* **14**, 2531 (2002).
9. H. Jaeger, H. Haas, Harnessing nonlinearity: Predicting chaotic systems and saving energy in wireless communication. *Science* **304**, 78 (2004).
10. K. Nakajima, I. Fischer, *Reservoir Computing—Theory, Physical Implementations, and Applications* (Springer, Singapore, 2021).
11. K. Nakajima, Physical reservoir computing—an introductory perspective. *Jpn. J. Appl. Phys.* **59**, 060501 (2020).
12. H. Hauser, A. J. Ijspeert, R. M. Fu¨chslin, R. Pfeifer, W. Maass, Towards a theoretical foundation for morphological computation with compliant bodies. *Biol. Cybern.* **105**, 355 (2011).
13. K. Caluwaerts, M. D'Haene, D. Verstraeten, B. Schrauwen, Locomotion without a brain: physical reservoir computing in tensegrity structures. *Artif. Life* **19**, 35 (2013).
14. Q. Zhao, K. Nakajima, H. Sumioka, H. Hauser, R. Pfeifer, Spine dynamics as a computational resource in spine-driven quadruped locomotion, in *2013 IEEE/RSJ International Conference on Intelligent Robots and Systems* (2013), pp. 1445–1451.
15. Y. Horii, K. Inoue, S. Nishikawa, K. Nakajima, R. Niiyama, Y. Kuniyoshi., Physical reservoir computing in a soft swimming robot, in *ALIFE 2022: The 2022 Conference on Artificial Life* (2022), pp. 92.
16. F. Daerden, D. Lefeber, Pneumatic artificial muscles: actuators for robotics and automation. *Eur. J. Mech. Env. Engineer.* **47**, 11 (2002).
17. D. Trivedi, C. D. Rahn, W. M. Kier, I. D. Walker, Soft robotics: Bi- ological inspiration, state of the art, and future research. *Appl. Bionics Biomech.* **5**, 99 (2008).
18. H. Schulte. The characteristics of the mckibben artificial muscle, in *The Application of External Power in Proshetics and Orhotics* (1961), pp. 94–115.
19. M. Gavrilovic, M. Maric, Positional servo-mechanism activated by artificial muscles. *J. Med. Biol. Eng.* **7**, 77 (1969).
20. T. Kanno, D. Morisaki, R. Miyazaki, G. Endo, K. Kawashima, A walking assistive device with intention detection using back-driven pneumatic artificial muscles, in *2015 IEEE International Conference on Rehabilitation Robotics (ICORR)* (2015), pp. 565–570.
21. D. Trivedi, D. Dienno, C. D. Rahn, Optimal, model-based design of soft robotic manipulators. *J. Mech. Des*. **9,** 130 (2008).

22. R. Sakurai, M. Nishida, T. Jo, Y. Wakao, K. Nakajima, Durable pneumatic artificial muscles with electric conductivity for reliable physical reser- voir computing. *J. Robot. Mechatron.* **34**, 240 (2022).
23. H. Hayashi, T. Kawase, T. Miyazaki, M. Sogabe, Y. Nakajima, K. Kawashima, Online assistance control of a pneumatic gait assistive suit using physical reservoir computing exploiting air dynamics, in *2022 International Conference on Robotics and Automation (ICRA)* (2022), pp. 3245–3251.
24. M. Eder, F. Hisch, H. Hauser, Morphological computation-based control of a modular, pneumatically driven. *Adv. Robot.* **32**, 375 (2018).
25. S. Iqbal, X. Zang, Y. Zhu, J. Zhao, Bifurcations and chaos in passive dynamic walking: A review. *Rob. Auton. Syst.* **62**, 889 (2014).
26. R. Terajima, K. Inoue, S. Yonekura, K. Nakajima, Y. Kuniyoshi, Behavioral diversity generated from body–environment interactions in a simulated tensegrity robot. *IEEE Robot. Autom. Lett.* **7**, 1597 (2022).
27. A. Pitti, R. Niiyama, Y. Kuniyoshi, Creating and modulating rhythms by controlling the physics of the body. *Auton. Robots* **28**, 317 (2010).
28. S. Steingrube, M. Timme, F. Worgotter, P. Manoonpong, Self- organized adaptation of a simple neural circuit enables complex robot behaviour. *Nat. Phys.* **6**, 224 (2010).
29. X. Zang, S. Iqbal, Y. Zhu, X. Liu, J. Zhao, Applications of chaotic dynamics in robotics. *Int. J. Adv. Robot. Syst.* **13**, 60 (2016).
30. R. Tokunaga, S. Kajiwara, T. Matsumoto, Reconstructing bifurcation diagrams only from time-waveforms. *Phys. D: Nonlinear Phenom.* **79**, 348 (1994).
31. Y. Itoh, Y. Tada, M. Adachi, Reconstructing bifurcation diagrams with lyapunov exponents from only time-series data using an extreme learning machine. *Nonlinear Theory and Its Applications, IEICE* **8**, 2 (2017).
32. Y. Itoh, S. Uenohara, M. Adachi, T. Morie, K. Aihara, Reconstructing bifurcation diagrams only from time-series data generated by electronic circuits in discrete-time dynamical systems. *Chaos* **30**, 013128 (2020).
33. J. Z. Kim, Z. Lu, E. Nozari, G. J. Pappas, D. S. Bassett, Teaching recurrent neural networks to infer global temporal structure from local examples. *Nat. Mach. Intell.* **3**, 316 (2021).
34. M. Hara, H. Kokubu, Learning dynamics by reservoir computing (in memory of prof. pavol brunovskỳ). *J. Dyn. Differ. Equ.* 1–26 (2022).
35. R. Sakurai, M. Nishida, H. Sakurai, Y. Wakao, N. Akashi, Y. Kuniyoshi, Y. Minami, K. Nakajima, Emulating a sensor using soft mate- rial dynamics: A reservoir computing approach to pneumatic artificial muscle, in *2020 3rd IEEE International Conference on Soft Robotics (RoboSoft)* (2020), pp. 710–717.
36. K. Yamaguchi, J. Busfield, A. Thomas, Electrical and mechanical behavior of filled elastomers. i. the effect of strain. *J. Polym. Sci. B: Polym. Phys.* **41**, 2079 (2003).
37. L. Appeltant, M. C. Soriano, G. V. Sande, J. Danckaert, S. Massar, J. Dambre, B. Schrauwen, C. R Mirasso, I. Fischer, Information processing using a single dynamical node as complex system. *Nat. Commun.* **2**, 1 (2011).
38. K. Fujii, K. Nakajima, Harnessing disordered-ensemble quantum dynamics for machine learning. *Phys. Rev. Appl.* **8**, 024030 (2017).
39. J. Torrejon, M. Riou, F. A. Araujo, S. Tsunegi, G. Khalsa, D. Querlioz, P. Bortolotti, V. Cros, K. Yakushiji, Neuromorphic computing with nanoscale spintronic oscillators. *Nature* **547**, 428 (2017).

40. J. Dambre, D. Verstraeten, B. Schrauwen, S. Massar, Information processing capacity of dynamical systems. *Sci. Rep.* **2**, 1 (2012).
41. T. Kubota, H. Takahashi, K. Nakajima, Unifying framework for information processing in stochastically driven dynamical systems. *Phys. Rev. Res.* **3**, 043135 (2021).
42. H. Jaeger, Short term memory in echo state networks. *GMD-Report, GMD-Forschungszentrum Informationstechnik* **152** (2001).
43. H. Jaeger, The "echo state" approach to analysing and training recurrent neural networks-with an erratum note, *German National Research Center for Information Technology GMD Technical Report* **148**, 13 (2001).
44. K. Xing, Y. Wang, Q. Zhu, H. Zhou, Modeling and control of mckibben artificial muscle enhanced with echo state networks. *Control Eng. Pract.* **20**, 477 (2012).
45. Y. Cao, J. Huang, Neural-network-based nonlinear model predictive tracking control of a pneumatic muscle actuator-driven exoskeleton. *J. Autom. Sin.* **7**, 1478 (2020).
46. W. Sun, N. Akashi, Y. Kuniyoshi, K. Nakajima, Physics-informed reservoir computing with autonomously switching readouts: a case study in pneumatic artificial muscles, in *The 32nd 2021 International Symposium on Micro-NanoMechatronics and Human Science* (2021), pp. 1–6.
47. W. Sun, N. Akashi, Y. Kuniyoshi, K. Nakajima, Self-organization of physics-informed mechanisms in recurrent neural networks: A case study in pneumatic artificial muscles, in *2022 IEEE 5th International Conference on Soft Robotics (RoboSoft)* (2022), pp. 409–415.
48. W. Sun, N. Akashi, Y. Kuniyoshi, K. Nakajima, Physics-informed recurrent neural networks for soft pneumatic actuators. *IEEE Robot. Autom. Lett.* **7**, 6862 (2022).
49. A. J. Ijspeert, Central pattern generators for locomotion control in animals and robots: a review. *Neural Netw.* **21**, 642 (2008).
50. O. E. Rossler, An equation for continuous chaos. *Phys. Lett. A* **57**, 397 (1976).
51. I. Bendixson, Sur les courbes définies par des équations différentielles. *Acta Math.* **24**, 1 (1901).
52. H. Hauser, A. J. Ijspeert, R. M. Fuchslin, R. Pfeifer, W. Maass, The role of feedback in morphological computation with compliant bodies. *Biol. Cybern.* **106**, 595 (2012).
53. F. Takens. Detecting strange attractors in turbulence. (Springer-Verlag, New York, 1981).
54. T.-Y. Li, J. A. Yorke, Period three implies chaos. In *The theory of chaotic attractors* (Springer, New York, 2004).
55. I. Inoue, K. Nakajima, Y. Kuniyoshi, Designing Spontaneous behavioral switching via chaotic itinerancy. *Sci. Adv.* **6**, 46 (2020).
56. C.-P. Chou, B. Hannaford, Measurement and modeling of mckibben pneumatic artificial muscles. *IEEE Trans. Robot. Autom.* **12**, 90 (1996).
57. B. Tondu, Modelling of the mckibben artificial muscle: A review. *Journal of Intelligent Material Systems and Structures*. *J. Intell. Mater. Syst. Struct.* **23**, 225 (2012).
58. H. Chaoui, P. Sicard, W. Gueaieb, Ann-based adaptive control of robotic manipulators with friction and joint elasticity. *IEEE Trans. Ind. Electron.* **56**, 3174 (2009).
59. W. Huang, X. Huang, C. Majidi, M. K. Jawed, Dynamic simulation of articulated soft robots. *Nat. Commun.* **11**, 1 (2020).
60. J. Y. Loo, Z. Y. Ding, V. M. Baskaran, S. G. Nurzaman, C. P. Tan, Robust multimodal indirect sensing for soft robots via neural network-aided filter-based estimation. *Soft Robot.* (2021).
61. R. Pfeifer, C. Scheier. *Understanding intelligence* (MIT press, 2001)

62. T. V. Minh, T. Tjahjowidodo, H. Ramon, H. V. Brussel, Cascade position control of a single pneumatic artificial muscle–mass system with hysteresis compensation. *Mechatronics*, **3**, 20 (2010)
63. M. Hofer, R. D'Andrea, Design, Modeling and Control of a Soft Robotic Arm in *2018 IEEE/RSJ International Conference on Intelligent Robots and Systems (IROS)* (2018), pp. 1456–1463.
64. W. Huang, X. Huang, C. Majidi, M. K. Jawed, Dynamic simulation of articulated soft robots, *Nat. Comnum.* **11**, 1 (2020)
65. K. H. Meyer, C. Ferri, The Elasticity of Rubber. *Rubber, Chemistry and Technology* **8**, 319 (1935)
66. A. F. Atiya, A. G. Parlos, New results on recurrent network training: unifying the algorithms and accelerating convergence. *IEEE transactions on neural networks* **11**, 697 (2000).
67. H. Kubota, K. Yakushiji, A. Fukushima, S. Tamaru, M. Konoto, T. Nozaki, S. Ishibashi, T. Saruya, S. Yuasa, T. Taniguchi, Spin-torque oscillator based on magnetic tunnel junction with a perpendicularly magnetized free layer and in-plane magnetized polarizer. *Appl. Phys. Express* **6**, 103003 (2013)

**Acknowledgments:** The results were partially obtained from the Innovative AI Chips and Next-Generation Computing Technology Development project and Development of Next-Generation Computing Technologies/Exploration of Neuromorphic Dynamics towards Future Symbiotic Society project, which were commissioned by NEDO.

**Funding:** NA received support from JSPS KAKENHI (Grant Number JP22J01542).
KN received support from JSPS KAKENHI (Grant Number JP18H05472).
KN received support from JST CREST (Grant Number JPMJCR2014).

**Author contributions:**
Conceptualization: KN, NA
Methodology: NA, TJ, MN, RS, YW, KN
Investigation: NA, RS, KN
Visualization: NA, RS
Funding acquisition: KN, NA
Project administration: KN
Supervision: KN, YK
Writing – original draft: NA, KN
Writing – review & editing: KN, YK, TJ, MN, RS, YW

**Competing interests:** Authors declare that they have no competing interests.

**Data and materials availability:** All data are available in the main text or the Supplementary Materials.

**Figures:**

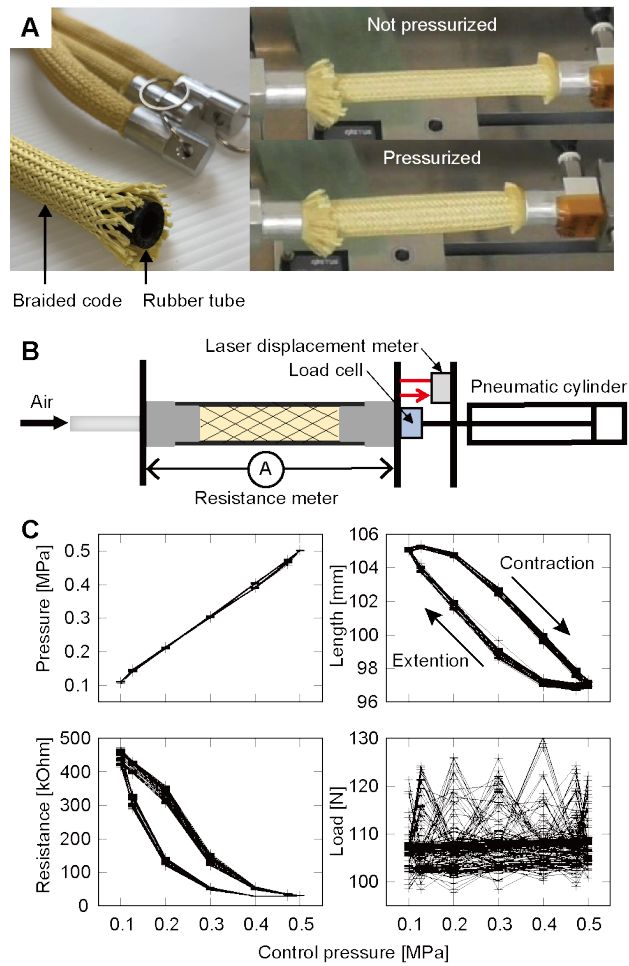

**Fig. 1. Pneumatic artificial muscles, measurement systems, and pneumatic artificial muscle dynamics.** (**A**) No pressurized and pressurized pneumatic artificial muscles. (**B**) Pneumatic artificial muscle measurement systems. (**C**) Sensor responses for a sinusoidal wave input.

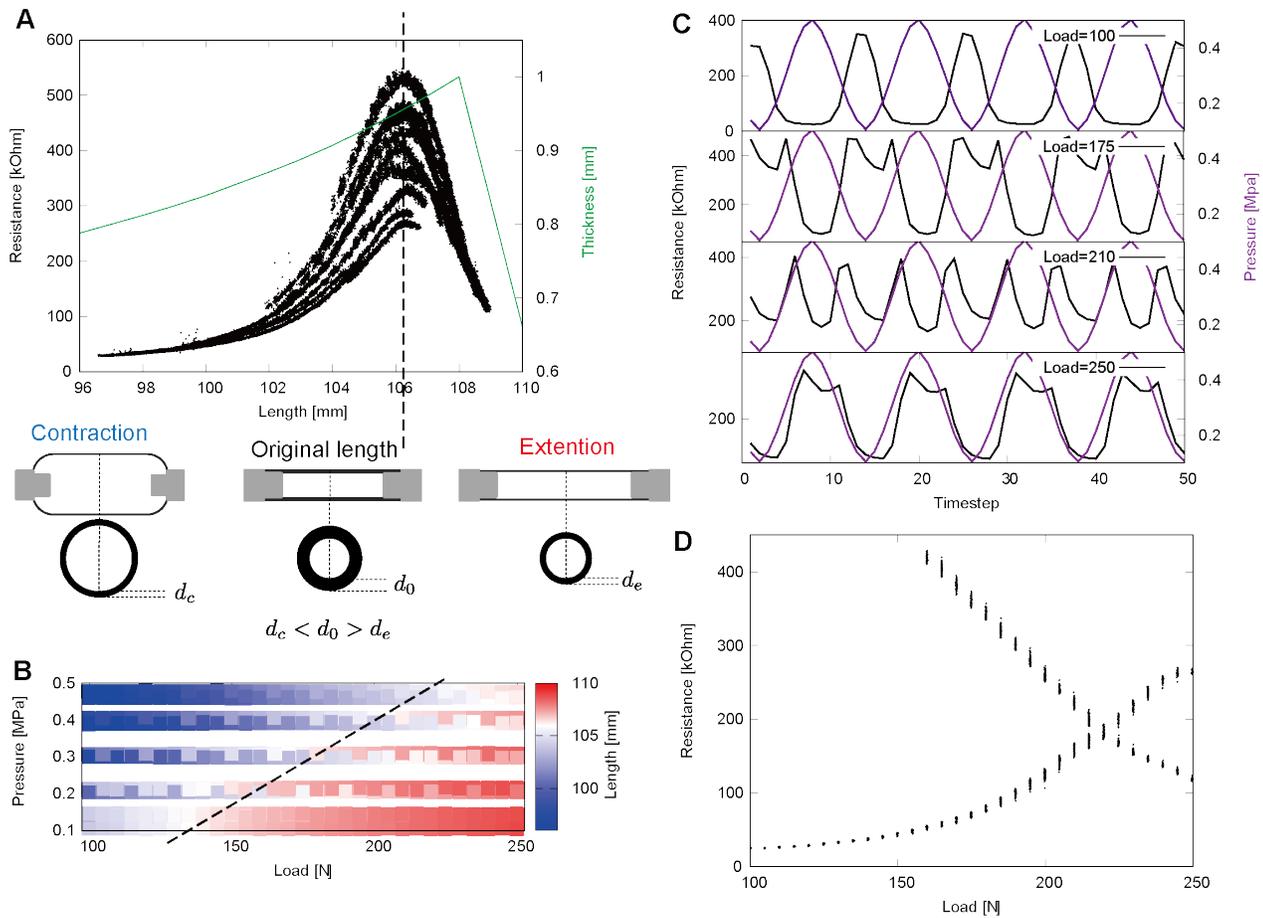

**Fig. 2. Intrinsic bifurcations of pneumatic artificial muscle.** (**A**) The top graph plots length versus resistance; the bottom figures depict the schematic illustration of the thickness change. (**B**) Color map of the length for control load and pressure. (**C**) Resistance and pressure time series in four load conditions. (**D**) Bifurcation diagram of control load versus local minimum resistance.

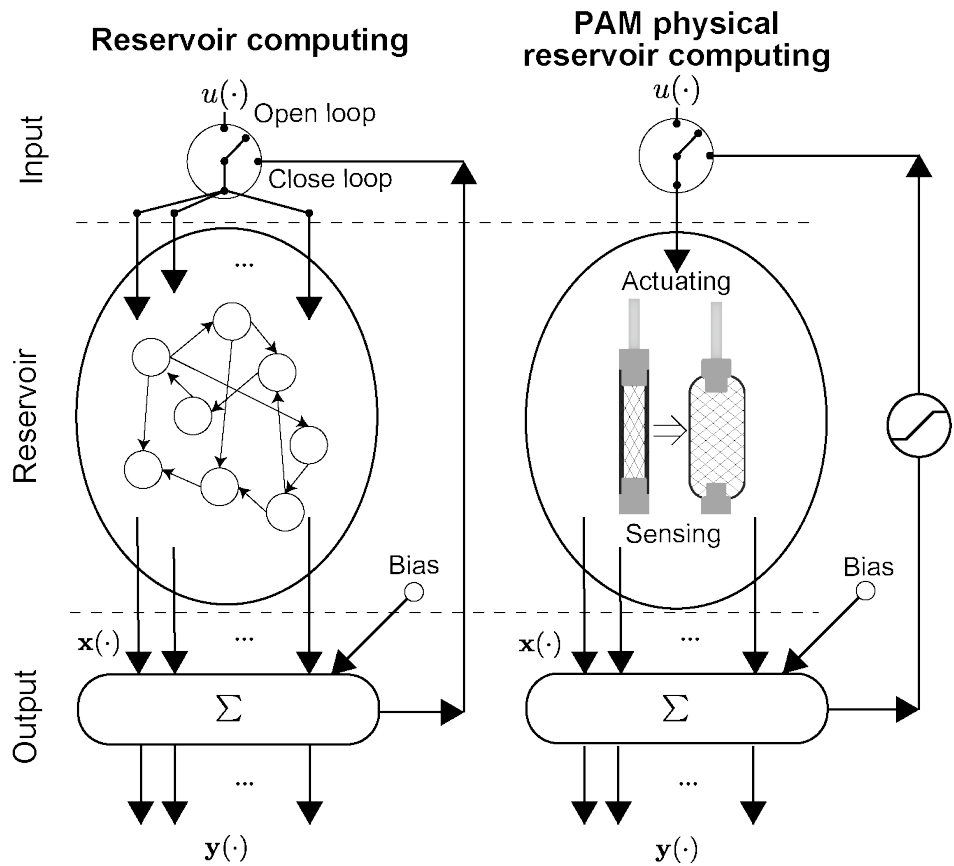

**Fig. 3. Schematics of reservoir computing. (A)** The right-hand side depicts a schematics of typical reservoir computing. The left-hand side depicts a schematics of Physical reservoir computing for pneumatic artificial muscle.

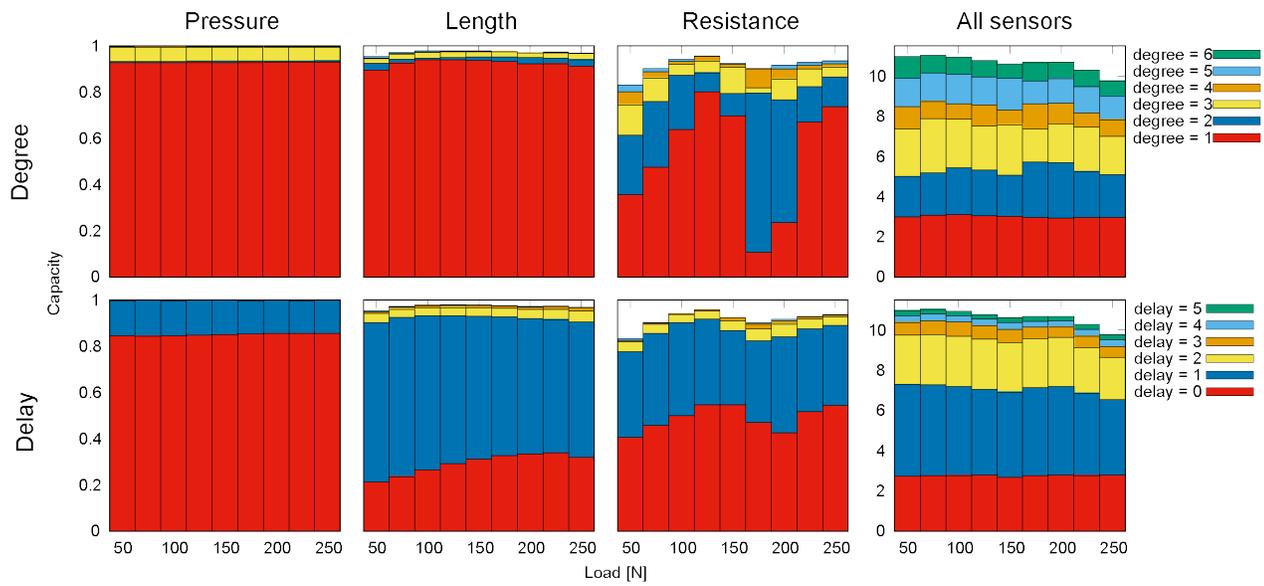

**Fig. 4. Information processing capacities of pneumatic artificial muscle sensory values.** The bars indicate the decompositions of the IPCs through the degree components and memory components (The method for the decomposition is provided in the Supplementary Materials 1).

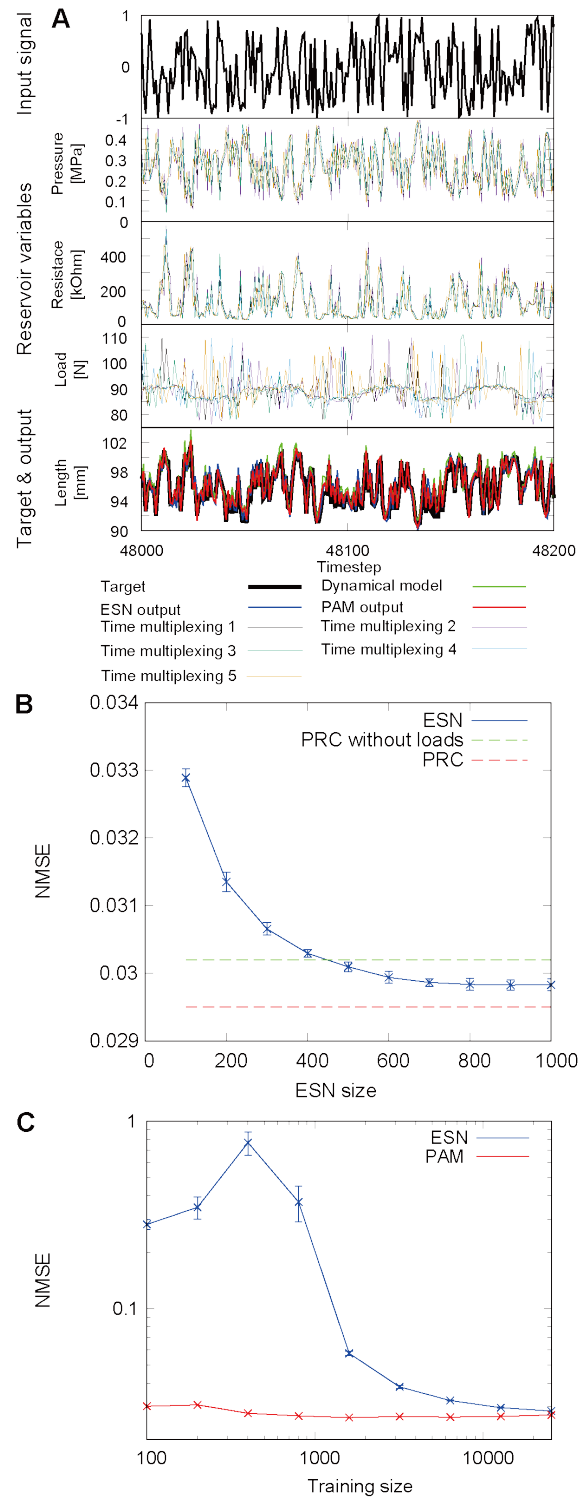

**Fig. 5. Open-loop length sensor emulation.** (**A**) Time series of the input, reservoir variables, output, and prediction in the length sensor emulation task. The red line is the prediction signal of physical reservoir computing for pneumatic artificial muscle. (**B**) Normalized mean squared error through the number of echo state network nodes. (**C**) Normalized mean squared error using the number of training data. Echo state network nodes: n = 600.

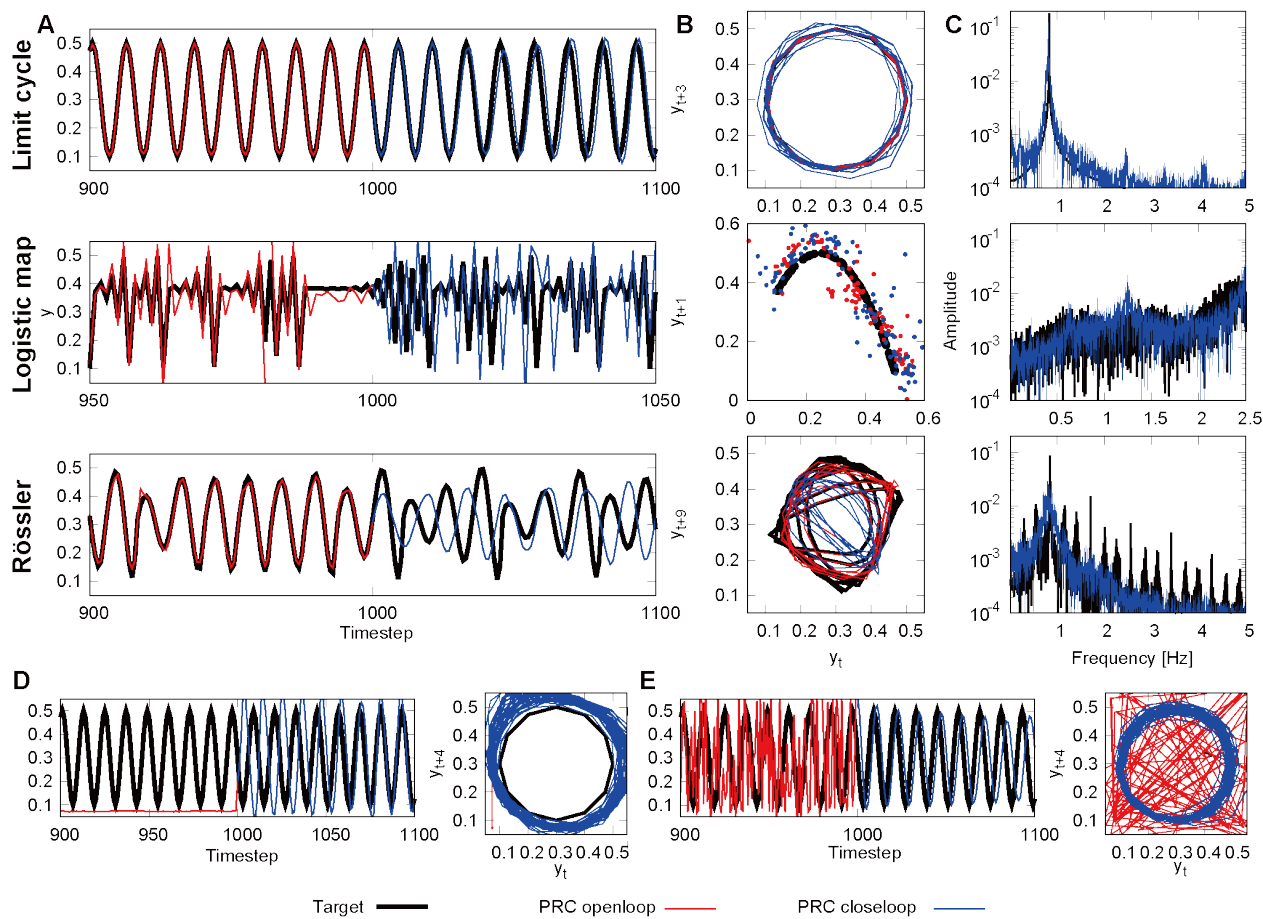

**Fig. 6. Results of the closed-loop at the attractor embedding.** (**A**) Time series of the target and physical reservoir computing output signals. (**B**) Attractors of the target and physical reservoir computing output. (**C**) Spectra of the target and physical reservoir computing output. (**D**) Time series and attractor in the limit cycle, embedding from zero inputs. (**E**) Time series and attractor in the limit cycle, embedding from a random input.

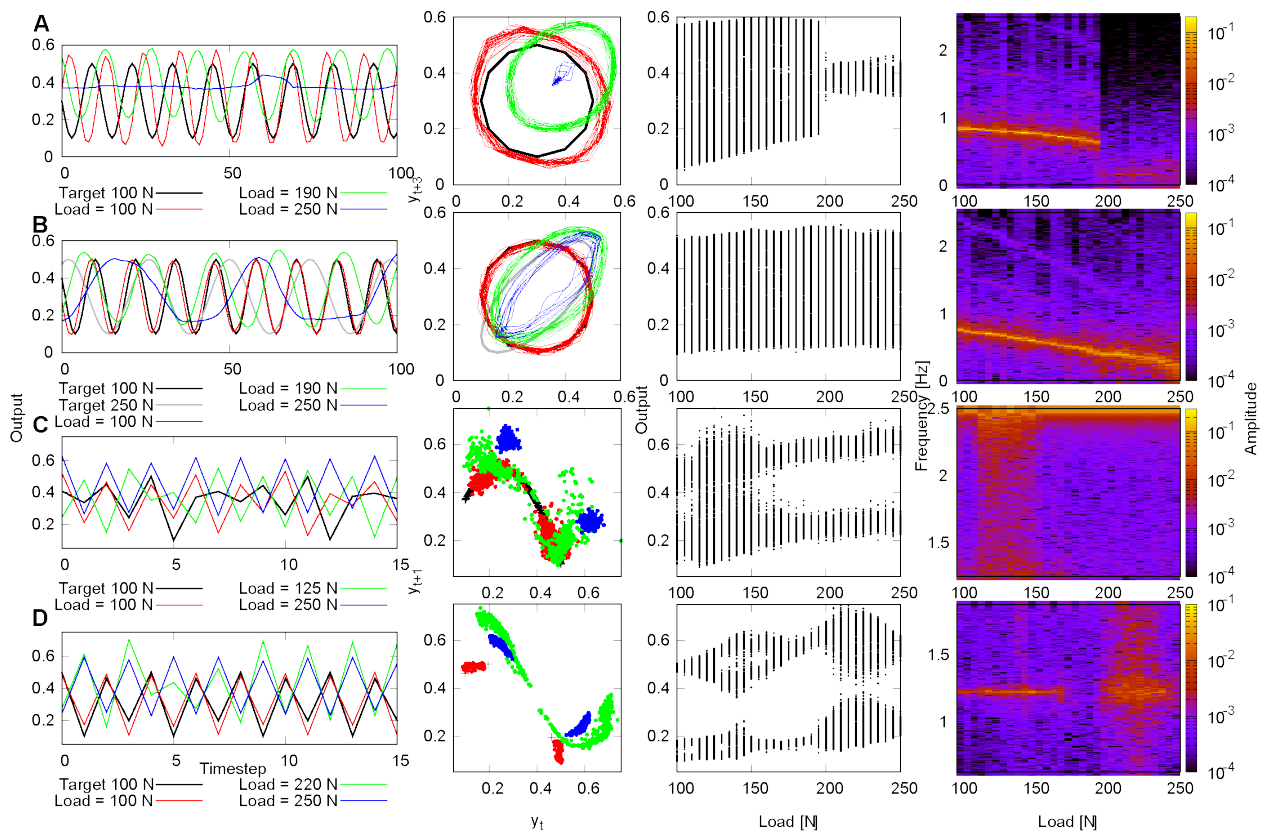

**Fig. 7. Bifurcation embedding using closed-loop control.** (**A**) Training the sinusoidal wave when the external load is 100 N. (**B**) Training sinusoidal waves with different periods when external loads are 100 N and 250 N. (**C**) Training the strange attractor of the logistic map when the external load is 100 N. (**D**) Training Period 4 trajectory when the external load is 100 N.

**Tables:**

Table 1. Major capacities of sensors.

| Target function | Pressure | Length | Resistance | Load | All sensors |
|---|---|---|---|---|---|
| IPC | 0.993 | 0.975 | 0.943 | 0.0037 | 9.861 |
| $\mathcal{P}_1(u_n)$ | 0.873 | 0.369 | 0.509 | 0.0037 | 0.957 |
| $\mathcal{P}_1(u_{n-1})$ | 0.089 | 0.527 | 0.223 | 0.000 | 0.996 |
| $\mathcal{P}_1(u_{n-2})$ | 0.000 | 0.026 | 0.013 | 0.000 | 0.769 |
| $\mathcal{P}_2(u_n)$ | 0.007 | 0.002 | 0.054 | 0.000 | 0.555 |
| $\mathcal{P}_1(u_n)\mathcal{P}_1(u_{n-1})$ | 0.006 | 0.012 | 0.095 | 0.000 | 0.340 |
| $\mathcal{P}_2(u_n)\mathcal{P}_1(u_{n-1})$ | 0.036 | 0.009 | 0.018 | 0.000 | 0.373 |